\documentclass{article}

\usepackage[preprint]{nips_2018}

\usepackage{graphicx}

\usepackage[utf8]{inputenc} 
\usepackage[T1]{fontenc}    
\usepackage{hyperref}       
\usepackage{url}            
\usepackage{booktabs}       
\usepackage{amsfonts}       
\usepackage{nicefrac}       
\usepackage{microtype}      
\usepackage{amsmath,amssymb}
\usepackage{mathtools}
\usepackage{enumitem}

\DeclareMathOperator*{\argmax}{arg\,max}

\title{Monte Carlo Tree Search for Asymmetric Trees}

\author{
  Thomas M. Moerland$^{\ast\dagger}$, Joost Broekens$^{\ast}$, Aske Plaat$^{\dagger}$ and Catholijn M. Jonker$^{\ast\dagger}$ \\
   \\
 $^\ast$Dep. of Computer Science, Delft University of Technology, The Netherlands \\ 
 $^\dagger$Dep. of Computer Science, Leiden University, The Netherlands
}

\begin{document}

\maketitle

\begin{abstract}
We present an extension of Monte Carlo Tree Search (MCTS) that strongly increases its efficiency for trees with asymmetry and/or loops. Asymmetric termination of search trees introduces a type of uncertainty for which the standard upper confidence bound (UCB) formula does not account. Our first algorithm (MCTS-T), which assumes a non-stochastic environment, backs-up tree structure uncertainty and leverages it for exploration in a modified UCB formula. Results show vastly improved efficiency in a well-known asymmetric domain in which MCTS performs arbitrarily bad. Next, we connect the ideas about asymmetric termination to the presence of loops in the tree, where the same state appears multiple times in a single trace. An extension to our algorithm (MCTS-T+), which in addition to non-stochasticity assumes full state observability, further increases search efficiency for domains with loops as well. Benchmark testing on a set of OpenAI Gym and Atari 2600 games indicates that our algorithms always perform better than or at least equivalent to standard MCTS, and could be first-choice tree search algorithms for non-stochastic, fully-observable environments.
\end{abstract} 

\section{Introduction} \label{sec_introduction}
Monte Carlo Tree Search (MCTS) \citep{coulom2006efficient} is a state-of-the-art algorithm in general game playing \citep{browne2012survey,chaslot2008monte}. The strength of MCTS is the use of statistical uncertainty to balance exploration versus exploitation \citep{munos2014bandits}, thereby effectively balancing breath and depth in the search tree. Probably the best known MCTS selection rule is Upper Confidence Bounds for Trees (UCT) \citep{kocsis2006bandit,cazenave2007parallelization}, which explores based on the Upper Confidence Bound (UCB) \citep{auer2002finite} of the mean action value estimate. More recently, MCTS also proved its benefit in combination with function approximation, either by generating targets for supervised learning \citep{guo2014deep}, or in combination with (self-play) reinforcement learning \citep{silver2016mastering,silver2017mastering}.

In this work we identify a fundamental deficit in the current MCTS algorithm. The problem is that MCTS cannot efficiently deal with asymmetric tree structure. Asymmetric tree structure occurs when the depth of the subtrees differs between the available actions at a state. MCTS does not take this uncertainty into account, because the information about the subtree structure is not backed-up in any way. In Section \ref{sec_asymmetric} we illustrate how MCTS can perform arbitrarily bad in domains with strong asymmetry, like the well-known RL task known as the `Chain' \citep{osband2016generalization}. 

This paper proposes a solution to this problem, by backing-up the uncertainty related to the tree structure ($\sigma_\tau$). By leveraging the tree structure uncertainty in the UCT exploration formula, we get a new algorithm (MCTS-T) that does efficiently solve asymmetric problems. Subsequently, we show that domain {\it loops},\footnote{It is important to discriminate a {\it loop} from a {\it transposition}. We provide details in the Discussion.} where the same state appears twice in the same trace, can be regarded as a special case of asymmetric termination. A simple extension (MCTS-T+) of the tree uncertainty back-ups further enhances search efficiency in domains with loops as well. Both our algorithms do assume non-stochastic environments, while MCTS-T+ additionally requires full state observability. Results on a set of OpenAI Gym tasks suggest that our algorithms could be first-choice when these assumptions are full-filled, especially since they are not harmful when the tree is actually symmetric without loops. 

The remainder of this paper is organized as follows. Section \ref{sec_preliminaries} provides essential preliminaries on Markov Decision Processes (MDP) and MCTS. In Section \ref{sec_asymmetric} we illustrate the asymmetric termination problem, introduce the MCTS-T algorithm, and show initial results on the Chain domain. Section \ref{sec_loops} identifies the problem of domain loops, extends the algorithm to MCTS-T+, and shows results on a Chain domain with loops. Section \ref{sec_experiments} tests our algorithms an a range of tasks from the OpenAI Gym repository, including a set of Atari 2600 games. We finish with a discussion (Section \ref{sec_discussion}) and conclusion (Section \ref{sec_conclusion}) of our work. 

\section{Preliminaries} \label{sec_preliminaries}

\subsection{Markov Decision Process}
We adopt a finite-horizon Markov Decision Process (MDP) \citep{sutton2018reinforcement} given by the tuple $\{\mathcal{S},\mathcal{A},f,\mathcal{R},\gamma,T\}$, where $\mathcal{S} \subseteq \mathbb{R}^{N_s}$ is a state set, $\mathcal{A} = \{1,2,..,N_a \}$ is a discrete action set, $f: \mathcal{S} \times \mathcal{A} \to \mathcal{S}$ denotes a deterministic transition function, $R: \mathcal{S} \times \mathcal{A} \to \mathbb{R}$ a (bounded) deterministic reward function, $\gamma \in (0,1]$ a discount parameter and $T$ the time horizon. At every time-step $t$ we observe a state $s^t \in \mathcal{S}$ and pick an action $a^t \in \mathcal{A}$, after which the environment returns a reward $r^t = \mathcal{R}(s^t,a^t)$ and next state $s^{t+1} = f(s^t,a^t)$. We act in the MDP according to a stochastic policy $\pi: \mathcal{S} \to P(\mathcal{A})$. Define the (policy-dependent)  state value $V^\pi(s_t) = \mathrm{E}_\pi[ \sum_{k=0}^T (\gamma)^k \cdot r^{t+k}]$ and state-action value $Q^\pi(s_t,a_t) = \mathrm{E}_\pi[ \sum_{k=0}^T (\gamma)^k \cdot r^{t+k}]$, respectively. Our goal is to find a policy $\pi$ that maximizes this cumulative, discounted sum of rewards.

\subsection{Monte Carlo Tree Search}
One approach to solving the MDP optimization problem is through planning. Before we perform an action $a^t$ in a state $s^t$, we get to expend some computational budget of forward simulation to find out which action is best. A particular successful class of planning algorithms are known as Monte Carlo Tree Search (MCTS). MCTS builds a search tree which it repeatedly traverses based on the upper confidence bound of the mean action value estimates. We will here introduce a variant of the PUCT algorithm \citep{rosin2011multi}, as this recently showed strong performance in the game of Go \citep{silver2017mastering}. Every state node $s$ in the tree has edges $(s,a)$ for each available action $a$. The edges store statistics $\{n(s,a),W(s,a),Q(s,a)\}$, where $n(s,a)$ is the visitation count, $W(s,a)$ the cumulative return over all roll-outs through $(s,a)$, and $Q(s,a) = W(s,a)/n(s,a)$ is the mean action value estimate. MCTS repeatedly performs four subroutines \citep{browne2012survey}:

\begin{enumerate}[leftmargin=1.0cm]
\item {\bf Select}
In the first stage of MCTS, we descend the tree from the root $s_0(:=s^t)$\footnote{We use superscript $s^t$ to index real environment states, and subscripts $s_d$ to index states at depth $d$ in a search tree. At every timestep $t$, the tree root $s_0:=s^t$, i.e. the current environment state becomes the tree root.} according to the {\it tree policy}. The tree policy selects actions according to statistics in the current tree:

\begin{equation}
\pi_{tree}(a|s) = \argmax_a \Bigg[ Q(s,a) + c \cdot \frac{\sqrt{n(s)}}{n(s,a)} \Bigg] \label{eq_tree_policy}
\end{equation}

where $n(s) = \sum_a n(s,a)$ is the total number of visits to state $s$, and $c \in \mathbb{R}^+$ is a constant that scales the amount the exploration/optimism. For an untried action ($n(s,a)=0$) the upper confidence bound is assumed to be $\infty$. The tree policy is followed until we either reach a terminal state or select an action we have not tried before. The tree policy naturally balances exploration versus exploitation, as it initially prefers all actions (due to low visitation count), but asymptotically only selects the optimal action(s). 

\item {\bf Expand} We next expand the tree with a new leaf state $s_L$ obtained from simulating the environment with the new action from the last state in the current tree. Subsequently, we initialize the child edges (actions) of the new leaf $s_L$ with statistics $\{n(s_L,a)=0,W(s_L,a)=0,Q(s_L,a)=0\} \forall a \in \mathcal{A}$.

\item {\bf Roll-out} We then require an estimate of the value $V(s_L)$ of the new leaf node, for which MCTS uses a random roll-out from $s_L$. We estimate $V(s_L)$ from the sum of rewards in the random roll-out: $\mathrm{R}(s_L) = \sum_{i=L}^D r(s_i,a_i)$, where $D$ denotes some termination depth. Note that the roll-out estimate can be improved by using a better roll-out policy or replaced by a learned value function, e.g. when learning off-line with neural networks as function approximator \citep{silver2016mastering,silver2017mastering}.

\item {\bf Back-up} Finally, we recursively back-up the results of the roll-out in the tree. Denote the current forward trace in the tree as $\{s_0,a_0,s_1, .. s_{L-1},a_{L-1},s_L\}$. Then, for each state-action edge $(s_i,a_i)$, $L > i \geq 0$, we recursively estimate the state-action value as

\begin{equation}
\mathrm{R}(s_i,a_i) = r(s_i,a_i) + \gamma \mathrm{R}(s_{i+1},a_{i+1}). \label{eq_value_onpolicy}
\end{equation} 

where $\mathrm{R}(s_L,a_L) := \mathrm{R}(s_L)$. We then increment $W(s_i,a_i)$ with the new estimate $\mathrm{R}(s_i,a_i)$, increment the visitation count $n(s_i,a_i)$ with 1, and set the mean estimate to $Q(s_i,a_i) = W(s_i,a_i)/n(s_i,a_i)$. We repeatedly apply this back-up one step higher in the tree until we reach the root node $s_0$. 

\end{enumerate}

This procedure is repeated until the overall MCTS trace budget $N_{\text{trace}}$ is reached. Finally, the real environment action $a^t$ is picked based on the highest visitation count at the root $s_0(=s^t)$: 

\begin{equation}
\pi(s^t) = \argmax_{a \in \mathcal{A}} n(s_0,a) \label{eq_final_policy}
\end{equation}

\section{Asymmetric Termination} \label{sec_asymmetric}
We now identify the problems of MCTS with asymmetric tree structure. MCTS uses the local uncertainty (based on the counts $n(s,a)$) to forward traverse the tree, see Eq. \ref{eq_tree_policy}. Thereby, the actions at the root node will have the lowest uncertainty, as they will have the highest number of visits. However, there is actually a `backward' component to uncertainty. We actually become completely certain about the value of an action when we have enumerated the entire subtree below that action.\footnote{Note that we assume deterministic environments in this paper. See the Discussion as well.} However, the standard MCTS visitation count does not contain this information, because it is not backed-up in any way. 

\begin{figure}[t]
  \centering
      \includegraphics[width=1.0\textwidth]{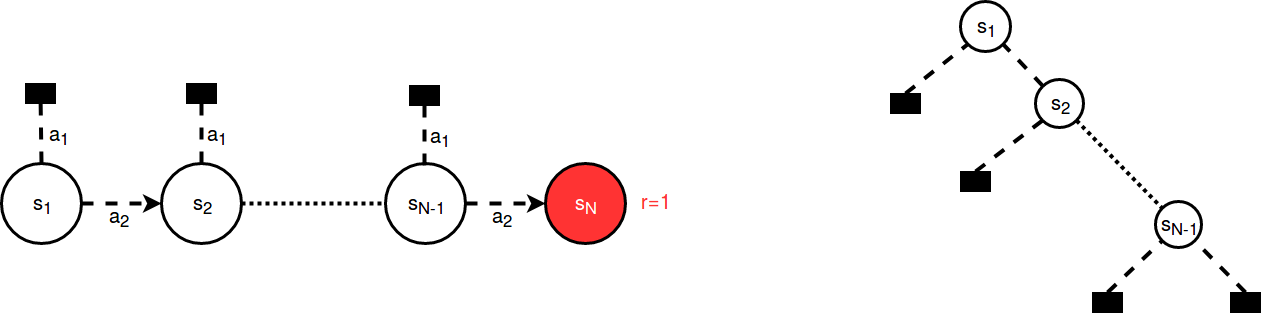}
  \caption{{\bf Left:} Chain domain. {\bf Right:} Search tree of the Chain domain.}
    \label{fig_asymmetric_chain}
\end{figure}

This termination uncertainty especially becomes useful when the underlying domain has an asymmetric tree structure. To illustrate this setting (and the suboptimal performance of MCTS) we consider a well-known RL domain: the Chain (Figure \ref{fig_asymmetric_chain}, left) \citep{osband2016generalization}. The Chain can be interpreted as a long, narrow path that we need to walk all the way to the end. At each timestep we have two actions available. One action always terminates the domain with a reward of 0. The other action moves the player one state further in the chain, but also gives a reward of 0. The only non-zero reward is received when the agent walks all the way along the chain up to depth $N$. The right of Fig. \ref{fig_asymmetric_chain} displays the search tree of the Chain domain. When visualizing the domain as a tree, we can clearly see the asymmetry. The domain is much deeper in one direction. Actually, the number of unique traces in this domain equals only $N+1$ for a Chain of length $N$, and exhaustive search solves the domain in $O(N)$ time complexity. 

Surprisingly, MCTS suffers a much higher sample complexity in this task. The problem is that for higher chain lengths, the MCTS roll-outs are unlikely to sample the full chain correctly and find the sparse reward. Therefore, both arms at the root will appear to be returning the same pay-off (of 0 in this case), and MCTS will equally spread its traces at the root node. This problem recursively appears at states further in the Chain. Therefore, for longer chains the time complexity of MCTS approaches the exponential $O(2^N)$, which is poor for a problem that actually has linear time complexity. The underlying problem is that MCTS does not back-up the fact that one of both arms at the root has been completely enumerated (which already happens after 1 trace), while the other direction still has unexplored traces.

\subsection{MCTS with Tree Uncertainty Back-up (MCTS-T)}
We now extend the MCTS algorithm to back-up and utilize the {\it uncertainty due to the tree structure} $\sigma_\tau(s) \in [0,1]$. For each state in the tree, we will estimate and recursively back-up $\sigma_\tau$, where  $\sigma_\tau(s)=1$ indicates a completely unexplored subtree below $s$, while $\sigma_\tau(s)=0$ indicates a fully enumerated subtree. We therefore define the $\sigma_\tau(s_L)$ of a newly expanded leaf state $s_L$ as:

\begin{equation}
\sigma_\tau(s_L) = \begin{cases}
0 &, \text{if }s_L\text{ is terminal} \\
1 &, \text{otherwise.}	\nonumber
\end{cases}
\end{equation}

We next recursively back-up $\sigma_\tau$ (the tree uncertainty) to previous states in the search tree, i.e., we update $\sigma_\tau(s_i)$ from the uncertainties of its successors $\sigma_\tau(s_{i+1})$. We could use a uniform policy for this back-up, but one of the strengths of MCTS is that it gradually starts to prefer (i.e., more strongly weigh) the outcomes of good arms. We therefore weigh the $\sigma_\tau$ back-ups by the empirical MCTS counts. Moreover, if an action has not been tried yet (and we therefore lack an estimate of $\sigma_\tau$), then we initialize the action as if tried once and with maximum uncertainty (the most conservative estimate). Defining 

\begin{figure}[t]
  \centering
      \includegraphics[width=1.0\textwidth]{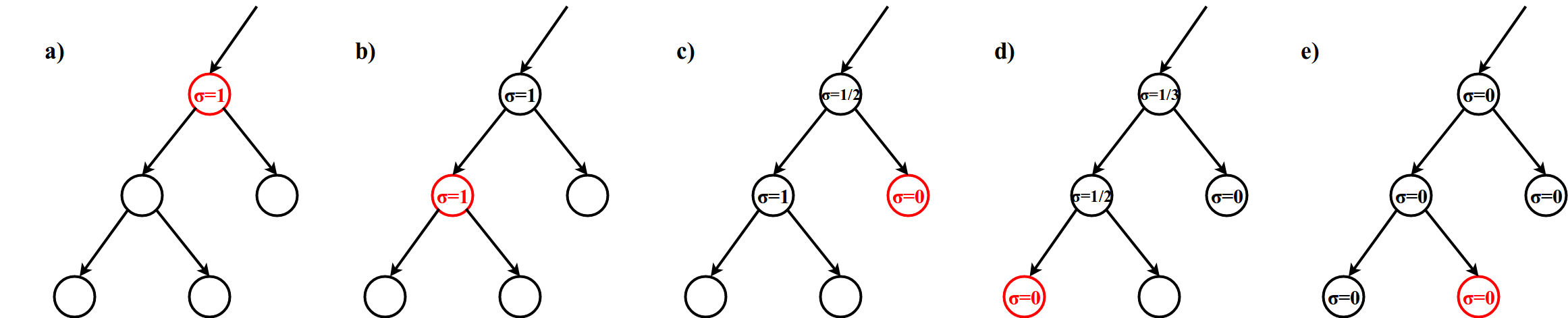}
  \caption{Process of $\sigma_\tau$ back-ups. Graphs a-e display subsequent estimates and back-ups of $\sigma_\tau$. In a) and b) we arrive at a non-terminal leaf node, of which the $\sigma_\tau$ automatically becomes 1. In the next subtree visit (c), we encounter a terminal leaf, and the uncertainty about the subtree at the subtree root decreases to $\frac{1}{2}$. In d) we encounter another terminal leaf. Because the back-ups are on-policy, we now estimate the root uncertainty as $\sigma_\tau = \frac{ (2 \cdot \frac{1}{2}) + (1 \cdot 0)}{2+1} = \frac{1}{3}$ (Eq. \ref{eq_sigma_backup}). Finally, at e) we enumerated the entire sub-tree, and the tree structure uncertainty at the subtree root is reduced to 0.}
    \label{fig_sigma_tree_backup}
\end{figure}

\begin{minipage}{.5\linewidth}
\begin{equation}
m(s,a) = \begin{cases}
n(s,a) &, \text{if } n(s,a) \geq 1 \\
1 &, \text{otherwise}	\nonumber
\end{cases}
\end{equation}
\end{minipage}
\begin{minipage}{.5\linewidth}
\begin{equation}
\sigma^\star_\tau(s') = \begin{cases}
\sigma_\tau(s') &, \text{if } n(s,a) \geq 1 \\
1 &, \text{otherwise,}
\end{cases}
\end{equation}
\end{minipage}

then the weighed $\sigma_\tau$ backup is 

\begin{equation}
\sigma_\tau(s) = \frac{\sum_{a} m(s,a) \cdot \sigma^\star_\tau(s')}{\sum_{a} m(s,a)} \label{eq_sigma_backup}
\end{equation}

for $s' = f(s,a)$ given by the deterministic environment dynamics. This back-up process is illustrated in Figure \ref{fig_sigma_tree_backup}.

\paragraph{Modified select step} Small $\sigma_\tau$ reduces our need to visit that subtree again for exploration, as we already (largely) know what will happen there. We therefore modify our tree policy at node $s$ to:

\begin{equation}
\pi_{tree}(s) = \argmax_a \Big [ Q(s,a) + c \cdot \sigma_\tau(s') \cdot \frac{\sqrt{n(s)}}{n(s,a)}  \Big ] \label{eq_ucb_sigma}
\end{equation}

for $s' = f(s,a)$ the successor state of action $a$ in $s$. The introduction of $\sigma_\tau$ acts as a prior on the upper confidence bound, reducing exploration pressure on those arms of which we have (largely) enumerated the subtree. 

\paragraph{Value back-up} MCTS uses {\it on-policy} value back-ups, i.e., the value of a state-action is estimated as the mean of all traces that passed through it. However, the $\sigma_\tau$ mechanism has introduced additional exploration pressure on deeper subtrees in the forward pass. In the value back-up, we ideally ignore our optimism of the forward pass (if necessary).\footnote{This problem disappears for number of traces $N \to \infty$, as the policy eventually becomes greedy on the optimal action. However, in practice we have to deal with finite computational budgets per step.} In reinforcement learning terminology, we ideally require an {\it off-policy} back-up. To stay as close to the original MCTS algorithm (which we use as our baseline comparison), we will use the standard UCT tree policy counts (without $\sigma_\tau$) as the empirical back-up policy. Define this separate set of `backward counts' $b(s,a)$ as obtained from Eq.\ref{eq_tree_policy}, then our value back-up is 

\begin{equation}
Q(s,a) = \frac{\sum_{a'} b(s',a') \cdot Q(s',a')}{\sum_{a'} b(s',a')}. \label{eq_value_backup}
\end{equation}

This requires calculating the UCB formula twice per timestep, but this increase is generally negligible compared to the computational time required for the environment simulations. Finally, because the forward counts are inflated in the direction of deeper subtrees, we can no longer use the counts at the root node for the final decision (Eq. \ref{eq_final_policy}) either. Instead, we base our decision on the highest mean action value among the actions at the root $s_0(=s^t)$:

\begin{equation}
\pi(s^t) = \argmax_{a \in \mathcal{A}} Q(s_0,a)
\end{equation}

This completes our proposed algorithm, which we call MCTS-T (MCTS with tree uncertainty).

\begin{figure}[b]
  \centering
      \includegraphics[width=1.0\textwidth]{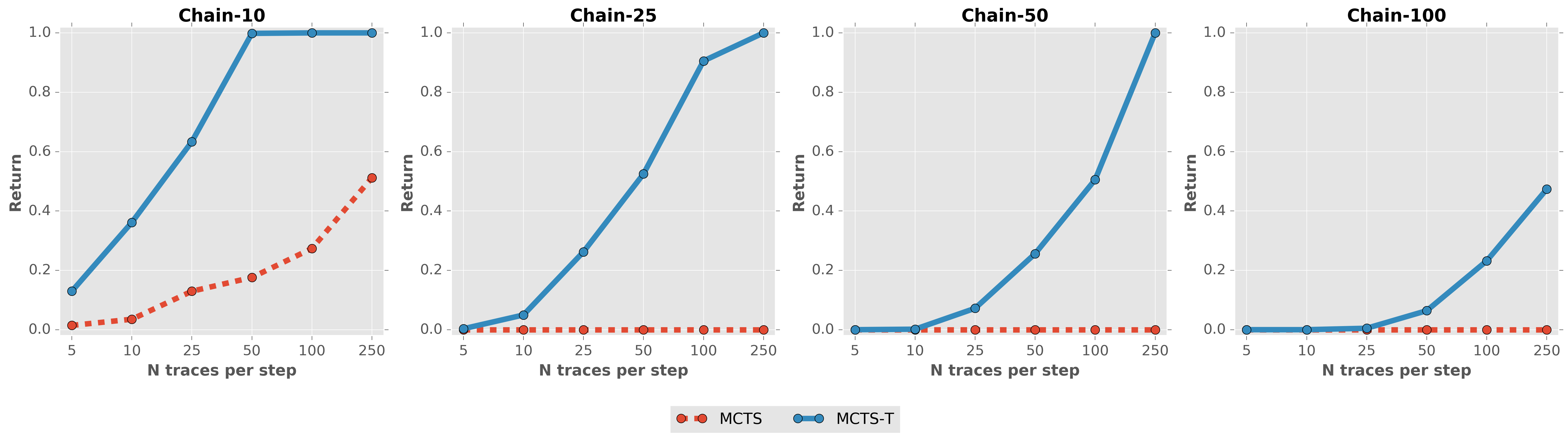}
  \caption{Comparison of vanilla MCTS (red) versus MCTS-T (blue) on the Chain domain of various lengths (progressing horizontally over the plots). Each plot displays computational budget per timestep (x-axis) versus average return per episode (y-axis). Results averaged over 25 episodes.}
    \label{fig_results_chain}
\end{figure}

\subsection{Results on Chain}
Figure \ref{fig_results_chain} shows the performance of MCTS versus MCTS-T on the Chain (Fig. \ref{fig_asymmetric_chain}). Plots progress horizontally for longer lengths of the Chain, i.e., stronger asymmetry and therefore a more difficult exploration challenge. In the short Chain of length 10 (Fig. \ref{fig_results_chain}, left), we see that both algorithms do learn, although MCTS-T is already more efficient. For the deeper chains of length 25, 50 and 100, we see that MCTS does not learn at all (as it starts to suffers an exponential time complexity). In contrast, MCTS-T does perform well in longer domains. 

\section{Loops} \label{sec_loops}

\begin{figure}[b]
  \centering
      \includegraphics[width=1.0\textwidth]{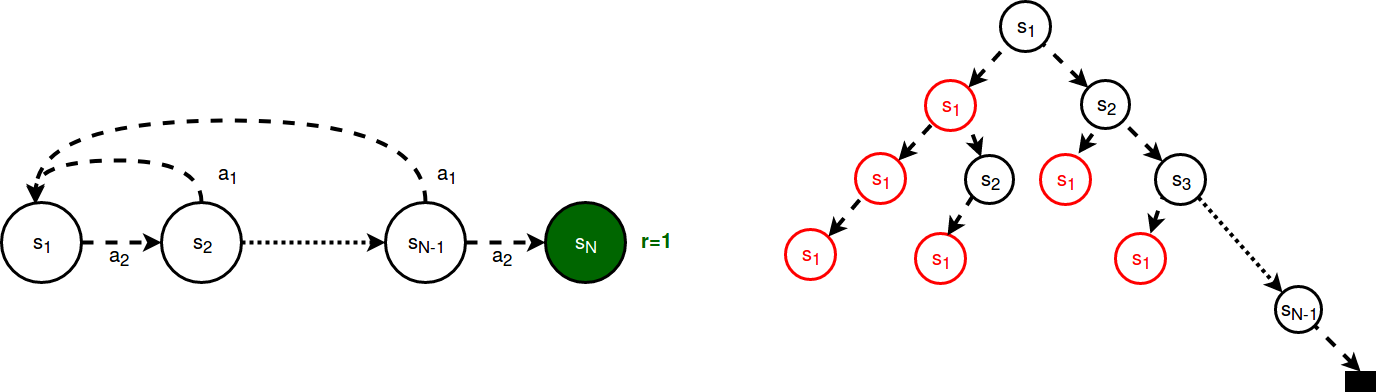}
  \caption{\small {\bf Left:} Chain domain with loops/cycles. {\bf Right:} Search tree of the cyclic Chain domain. Red nodes indicate a loop, i.e., the repetition of a state which already occurred in the trace above it.}
    \label{fig_cyclic_chain}
\end{figure}

We will now generalize the ideas about tree asymmetry to the presence of {\it loops} in the domain. A loop occurs when the same state appears twice in a single trace.\footnote{Loops actually occur in many RL tasks, for example navigation tasks or Atari 2600 games, where we may step in one direction and (approximately) return a few steps later. Such loops waste a lot of computational effort.} We will conceptually illustrate this in a variant of the Chain where the `wrong' action at each timestep returns the agent to state $s_1$, i.e., without terminating the episode (Figure \ref{fig_cyclic_chain}, left). Figure \ref{fig_cyclic_chain}, right displays the unfolded tree structure of this domain. This tree clearly shows that state $s_1$ repeatedly occurs in nearly all traces. Below each repetition, the entire root tree essentially repeats itself. The problem is that {\it everything} that might be learned below such a repetition, should actually be learned at the first occurrence of the state. However, standard MCTS will keep on expanding these nodes as if they are truly novel. 
 
\subsection{MCTS-T+: blocking loops.}
The solution of this problem is a natural extension of the $\sigma_\tau$ mechanism introduced in the previous section. Most importantly, when we encounter a duplicate state $s^\circ$ in the trace, we set $\sigma_\tau(s^\circ)=0$. Thereby, we pretend as if the looped state is terminal with respect to its tree uncertainty, because there is nothing novel to explore beyond that state (i.e., other actions should be explored at the first occurrence of the looped state). The value/roll-out estimate of the duplicate state $\mathrm{R}(s^\circ)$ technically depends on the sum of reward in the loop $\mathrm{S}^\circ = \sum_{s,a \in g} r(s,a)$, where $g =\{s^\diamond,..,s^\circ\}$ specifies the subset of the trace containing the loop (i.e., $s^\diamond = s^\circ$). For infinite time-horizon problems, we can theoretically repeat the loop forever, and therefore:

\begin{equation}
R(s^\circ) = \begin{cases}
\infty &, \text{if } \mathrm{S}^\circ \geq 0 \\
-\infty &, \text{if } \mathrm{S}^\circ \leq 0 \\
0 &, \text{if } \mathrm{S}^\circ = 0
\end{cases}
\end{equation}

For finite horizon problems we need to account for the number of remaining steps that we can repeat the loop. However, most frequently loops with a net positive or negative return are actually a domain artifact, as the solution of a (real-world) sequential decision making task is usually not to walk the same loop forever. 

We can efficiently check for a domain loop during the back-up phase of a trace. For each trace, we compare the expanded state $s_L$ to every state in the trace above it. When the norm between the expanded state $s_L$ and a previous state $s_i$, $0\leq i<L$, is smaller than some threshold $\eta$, we set $\sigma_\tau(s_L^\circ)=0$ and $R(s_L^\circ)$ as indicated above. Besides that, all the methodology from the previous section applies. Checking for a domain loop does incur an extra computational burden, although its effect is again relatively small compared to the cost of environment simulations in more complex domains (like Atari 2600 games). We call the extended algorithm with loop detection MCTS-T+. 

\subsection{Results on Chain with loops.}
\begin{figure}[t]
  \centering
      \includegraphics[width=1.0\textwidth]{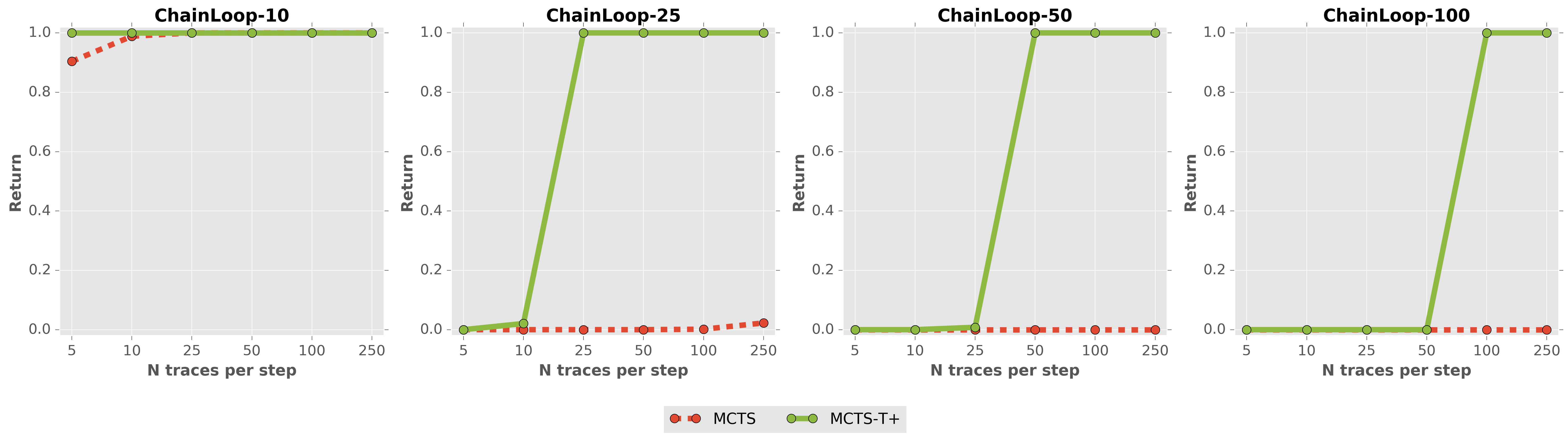}
  \caption{Comparison of MCTS (red) versus MCTS-T+ (green). MCTS-T+ uses tree uncertainty and loop blocking. Chain length progresses over the horizon plots. Results averaged over 25 episodes.}
    \label{fig_results_chain_loop}
\end{figure}

We illustrate the performance of MCTS-T+ on the Chain with loops (Figure \ref{fig_cyclic_chain}). The results are shown in Figure \ref{fig_results_chain_loop}. We observe a similar pattern as in the previous section, where MCTS only (partially) solves the shorter chains, but does not solve the longer chains at all. In contrast, MCTS-T+ does efficiently solve the longer chains as well. Note that MCTS-T (with loop detection) does not solve this problem either (curves not shown), as the loops prevent any termination, and therefore all $\sigma_\tau$ estimates stay at 1.

\section{Gym Experiments} \label{sec_experiments}
\begin{figure}[b]
  \centering
      \includegraphics[width=0.5\textwidth]{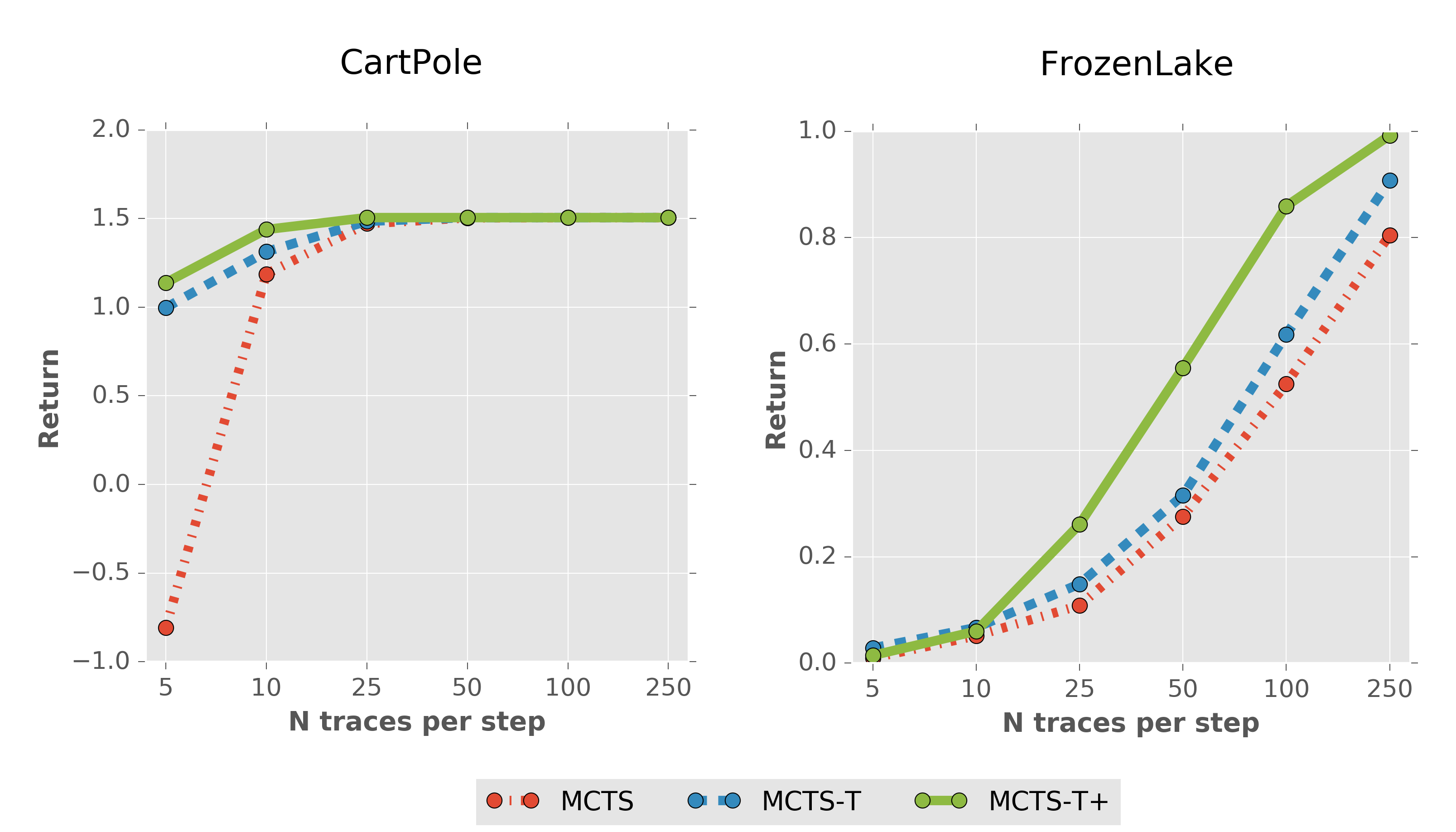}
  \caption{Learning curves on CartPole and FrozenLake. CartPole rewards are 0.005 for every timestep that the pole does not fall over, and -1 when the pole falls (episode terminates). Compared to the default Gym implementation, we use a FrozenLake variant without stochasticity. Episodes last 400 steps.}
    \label{fig_cartpole}
\end{figure}

\begin{figure}[t]
  \centering
      \includegraphics[width=1.0\textwidth]{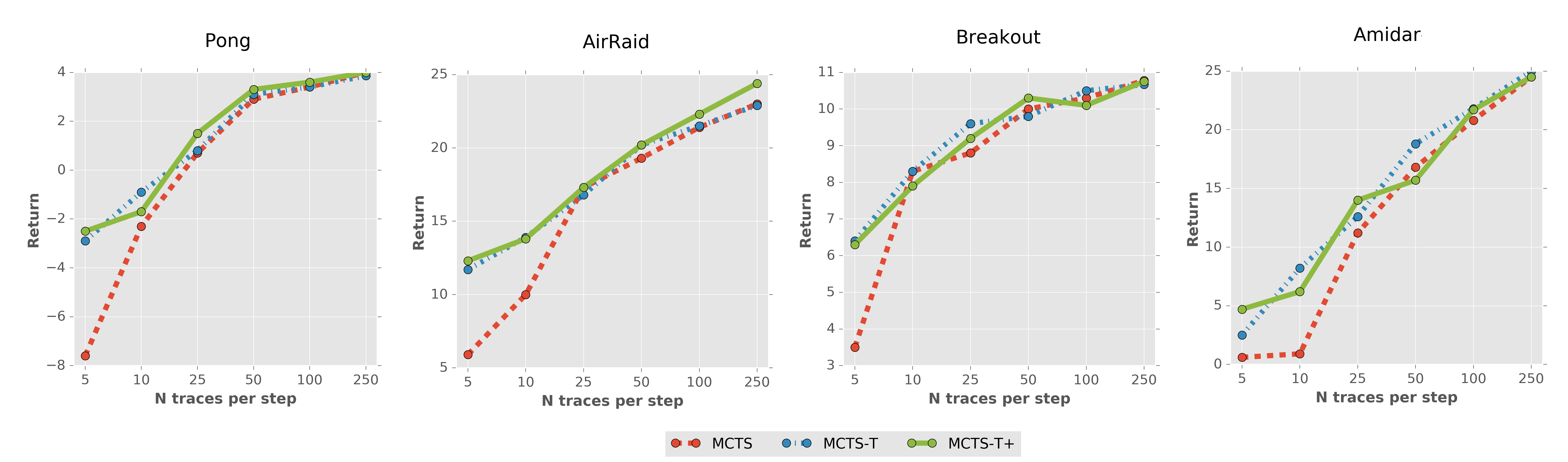}
  \caption{Learning curves on several Atari 2600 games. All domains clip rewards to $[-1,1]$. Each episode simulates 400 steps with a frameskip of 3.}
    \label{fig_atari}
\end{figure}

The previous experiments on the Chain are of course extreme cases of asymmetry and looping, where the tree uncertainty and loop blocking have most potential to be beneficial. We now experiment with our algorithms on a set of tasks from the OpenAI Gym repository, to verify whether these ideas are still beneficial in a general set of tasks. Figure \ref{fig_cartpole} shows the learning performance on CartPole and FrozenLake, respectively. We see that MCTS-T and MCTS-T+ consistently outperform MCTS, especially for a small number of traces. This is probably due to the fact that MCTS-T(+) can more efficiently direct its traces. Figure \ref{fig_atari} shows the results of our algorithms on a subset of Atari 2600 games. Again, both MCTS-T and MCTS-T+ outperform MCTS for a smaller number of traces, and generally perform at least equally well when $N$ grows larger. Code to replicate experiments is available from \url{https://github.com/tmoer/mcts-t.git}. 

\section{Discussion} \label{sec_discussion}
There are two important assumptions that need to be fulfilled for the current work. First, the current work is not yet applicable to stochastic domains. The problem with stochastic domains is that we can never bring the uncertainty of an arm to 0 based on fully enumerating each unique trace. Therefore, we should probably not add $\sigma_\tau$ in the UCB formula like in the present paper, as this completely collapses the confidence interval when the subtree is enumerated. Nevertheless, tree structure uncertainty is still relevant in stochastic domains, and coming up with a different incorporation in the UCB equation would be an interesting direction for future work. Second, loop blocking (MCTS-T+) can only be applied with fully observable states. For example, if we detect a loop in a partially observable domain, then the true environment state may actually have changed, and it could be harmful to block the apparent loop. 

Standard MCTS has predominantly shown its success in two-player board games. These environments, like the game of Go, usually show a rather symmetric search tree without many loops. In contrast, many tasks in which MCTS has shown less impressive results, like robotic control, single-player video games and navigation tasks, do also exhibit more asymmetry and loops. For example, in Atari 2600 a variety of wrong moves may suddenly end the game, while navigation tasks frequently allow to step back to a previous state after a few moves. The present work may therefore extend the applicability of MCTS to robotics control, navigation and single player video games.

We want to stress that a loop is something different than a transposition \citep{plaat1996exploiting}. In tree search, a transposition happens when a particular state appears in the subtrees of multiple arms of a state. As an illustration, in Fig.\ref{fig_cyclic_chain} right, all the red $s_1$ states are loops (as state $s_1$ already appears in the same trace at the root). However, state $s_2$ shows a transposition, as it occurs both in the left (at depth 2) and right arm from the root state $s_1$.  Transpositions can for example be handled by off-line learning and generalization of a value function, which stores what we already learned about the transposition in a different arm.

Another open problem in tree search is determining the number of traces before making a decision in the real environment. In practice, the number of traces per search $N$ is usually fixed in advance and always fully expended. This implies that we spend our full budget even when we are very close to domain termination. In contrast, the $\sigma_\tau$ back-up provides us with additional information to prematurely stop the search if new traces can no longer provide useful information. A simple choice would be to stop the search when $\sigma_\tau=0$ for all root arms (which implies fully enumerated subtrees).

\section{Conclusion} \label{sec_conclusion}
This paper introduces two extensions to standard MCTS. For non-stochastic environments, MCTS-T improves MCTS efficiency by backing-up the uncertainty due to the structure of the tree ($\sigma_\tau$). Under the additional assumption of full state observability, MCTS-T+ further extends the algorithm to also efficiently deal with potential loops in the domain. Our results indicate that MCTS-T(+) significantly outperforms MCTS on domains with strong asymmetry and loops. Moreover, MCTS-T(+) also improves performance on domains with less asymmetry, and - importantly - never performs suboptimal compared to standard MCTS. Therefore, MCTS-T(+) could be a first-choice search algorithm for domains with frequent asymmetry, like robotic control, navigation and single-player video games. 

\clearpage
\bibliographystyle{apalike}
\bibliography{example}

\end{document}